\documentclass[10pt,twocolumn,letterpaper]{article}

\usepackage{iccv}
\usepackage{times}
\usepackage{epsfig}
\usepackage{graphicx}
\usepackage{amsmath}
\usepackage{amssymb}

\usepackage[pagebackref=true,breaklinks=true,letterpaper=true,colorlinks,bookmarks=false]{hyperref}

\iccvfinalcopy 

\def\iccvPaperID{7966} 

\ificcvfinal\fi

\begin{document}

\title{Real-time 3D Object Detection using Feature Map Flow}

\author{Youshaa Murhij\\
Moscow Institute of Physics and Technology\\
Dolgoprodny, Russia\\
{\tt\small yosha.morheg@phystech.edu}
\and
Dmitry Yudin\\
Moscow Institute of Physics and Technology\\
Dolgoprodny, Russia\\
{\tt\small yudin.da@mipt.ru}
}

\maketitle
\ificcvfinal\thispagestyle{empty}\fi

\begin{abstract}
In this paper, we present a real-time 3D detection approach considering time-spatial feature map aggregation from different time steps of deep neural model inference (named feature map flow, FMF).
Proposed approach improves the quality of 3D detection center-based baseline and provides real-time performance on the nuScenes and Waymo benchmark. Code is available at \url{https://github.com/YoushaaMurhij/FMFNet}
\end{abstract}

\section{Introduction}
3D LiDAR point clouds provide much more precise spatial details \cite{yang2019std} compared to RGB, RGB-D or stereo images  from cameras \cite{basharov2021real} but they come in unordered and sparse form. 
Proposed method includes a development of a neural network for road scene object detection based on a recurrent single-stage object detector using multi-dimensional feature maps generated from two adjacent 3D point clouds.

\begin{figure}[ht]
\begin{center}
\includegraphics[width=1.0\linewidth]{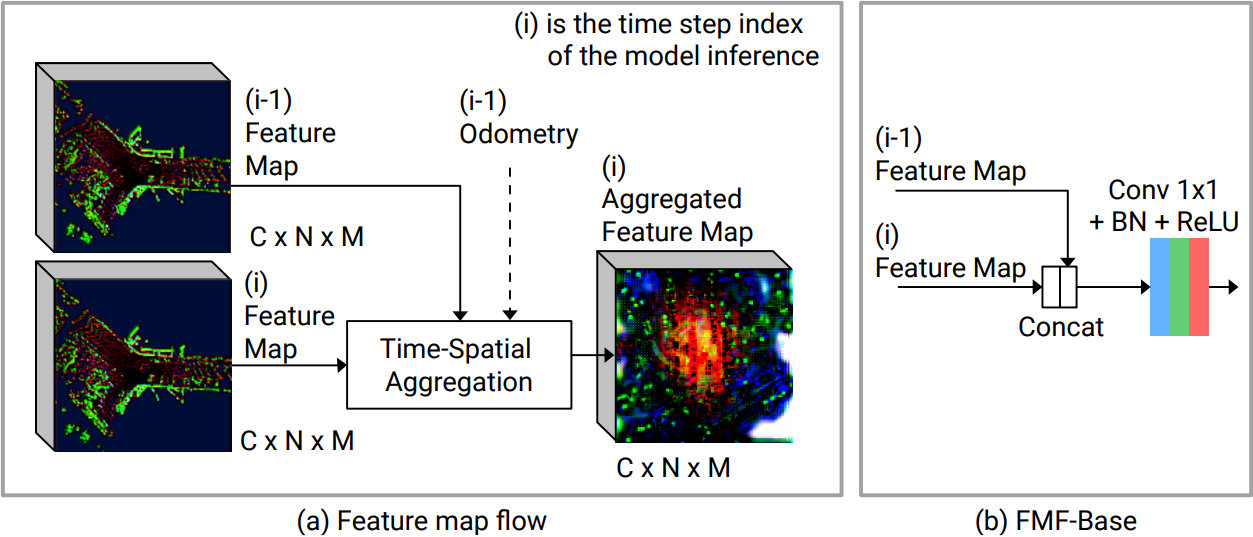}
\end{center}
   \caption{Proposed approach to time-spatial aggregation of two feature maps from different time steps of deep neural model inference (named feature map flow, FMF): (a) general FMF structure, odometry data is optional, (b) FMF-Base is concatenation with subsequent convolution, batch normalization and ReLU activation.}
\label{fig:fmf}
\end{figure}

Our contributions are the following: 

\begin{itemize}
    \item we proposed an approach of feature map time-spatial aggregation named feature map flow (FMF), see Figure~\ref{fig:fmf}; 
    \item we have demonstrated that FMF  improves the quality of detection of various basic state-of-the-art models and provides  real-time  performance  on  the nuScenes dataset and Waymo  benchmark.
    

    
\end{itemize}

\section{Related Work}
Processing point clouds in convolutional networks for detection task is computationally expensive. To enhance the processing efficiency, single stage approaches are used as in \cite{9157660} to downscale the 3D point cloud using lightweight Spatial-Semantic Feature Aggregation module to fuse high-level abstract semantic features and low-level spatial features for accurate predictions of bounding boxes and classification confidence. In other works \cite{Li20173DFC}, hand crafted voxel grid was used to extract voxel features. While VoxelNet model \cite{zhou2017voxelnet} removed the need of manual feature engineering for 3D point clouds and adopted a generic 3D detection network that unifies feature extraction and bounding box prediction into a single stage, end-to-end trainable deep network. The point cloud is divided to equally spaced 3D voxels and a group of points within each voxel are transformed into a unified feature representation through a voxel feature encoding (VFE) layer. 


Learning from unordered point sets was introduced in PointNets \cite{Qi2017PointNetDL} that was key network for PointPillars encoder \cite{8954311}, which utilizes PointNets to learn a representation of point clouds organized in vertical columns (pillars). In our work, PointPillars backbone is mainly used as a real-time detection model.

\begin{figure*}[t]
\begin{center}
\includegraphics[width=1.0\linewidth]{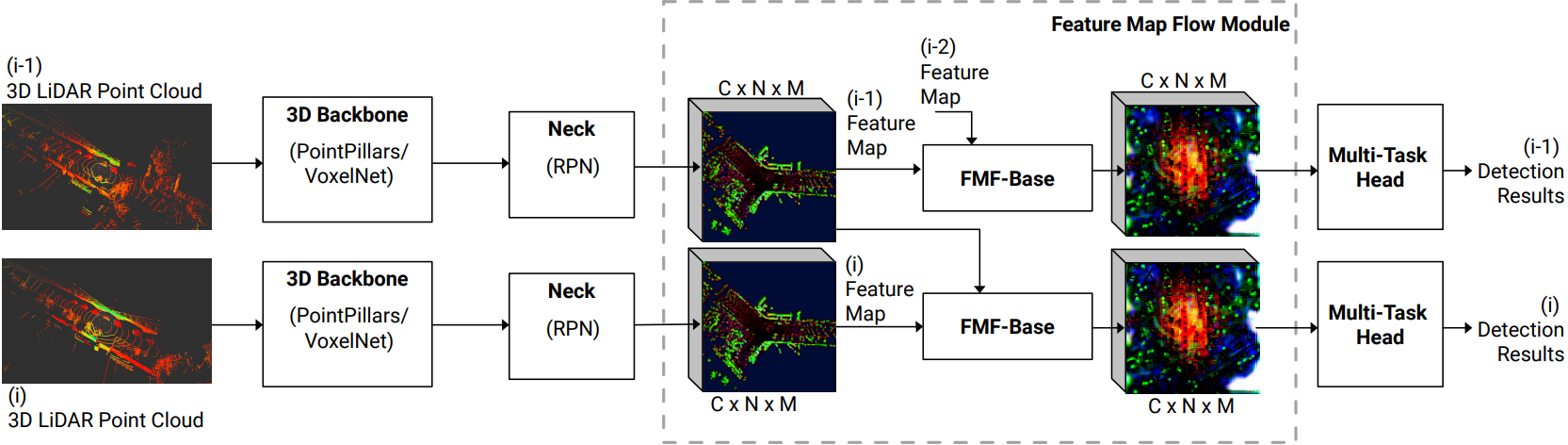}
\end{center}
   \caption{Architecture of the proposed method for 3D object detection with feature map flow module named FMFNet.}
\label{fig:FMFNet}
\end{figure*}

Multi-head network was applied in \cite{NEURIPS2018_432aca3a} and \cite{zhu2019classbalanced}, where sparse 3D convolution are utilized to extract rich semantic features, which are then fed into a class-balanced multi-head network to perform 3D object detection. To handle the severe class imbalance problem inherent in the autonomous driving scenarios, a class-balanced sampling and augmentation strategy  was designed to generate a more balanced data distribution. 


While center-based 3D object detection and tracking method \cite{yin2021centerbased} is designed based on CenterNet \cite{duan2019centernet} to represent, detect, and track 3D objects as points. CenterPoint framework \cite{yin2021centerbased}, first detects centers of objects using a keypoint detector and regresses to other attributes, including 3D size, 3D orientation, and velocity. In a second stage, CenterPoint refines these estimates using additional point features on the object.

\section{Method}

We study recurrent deep neural model for 3D object detection with architecture shown on Figure \ref{fig:FMFNet}.

Our model takes LiDAR point clouds as input and estimates oriented 3D boxes for objects on interest (cars, pedestrians, etc).

We use two common backbones (PointPillars \cite{8954311} and VoxelNet \cite{zhou2017voxelnet}) in our method for converting a point cloud to a sparse pseudoimage. 

Then, Neck block based on region proposal network (RPN) \cite{ren2016faster} processes the pseudo-image into high-level bird's eye view representation.

Proposed feature map flow module makes the time-spatial aggregation of these representations from subsequent model inference time steps. Its general structure is shown on Figure \ref{fig:fmf}(a).

The result of the aggregation is fed to the multi-task head, which allows us to find the two-dimensional coordinates of the centers of objects, estimate their height, 3D box size, orientation angle of the object on the plane. So, the problem of detection is reduced to the 2.5D case, which, however, quite effectively describes the behavior of three-dimensional objects in real traffic conditions.


Further in this section we will consider the work of the main module of the proposed approach in more detail.

\subsection{Feature Map Flow Module}
We propose a recurrent time-spatial aggregation of latent feature maps. The main idea behind the feature map flow block is to use the 2D feature map that was build in the previous detection and regression step before feeding the current generated map to the detection head.
Figure~\ref{fig:fmf}(a) shows detailed scheme of the proposed feature map flow module. In general, it can use the odometry data for the affine transformation of the feature map. But in this paper we are considering a simpler FMF-Base version, see Figure \ref{fig:fmf}(b).

In the FMF-Base module, we use both the concatenation and shared convolutional block with batch normalization and ReLU activation function that generates the final 2D feature map. In a later stage, this 2D map will be fed to the detection head module.

\subsection{Multi-task Head for 3D Object Detection}


We followed CenterPoint \cite{yin2021centerbased} method in our detection head implementation to generate heat map peak at the center location of the detected objects by producing a $K$-channel heat map $\hat{M}$, one channel for each of $K$ classes.

Projection of 3D centers of annotated bounding boxes into the map view is needed during training the model.
We use the Focal loss as in CenterNet \cite{duan2019centernet}.
Hence, 3D object detection turned in a keypoint estimation task as the raw point cloud converted to 2D grid to predict a $w \times h$ heat map $\hat{M} \in [0,1]^{w \times h \times K}$ for each class. The resulted heat map contains local maximums that represent the centers of the detected objects. We followed \cite{duan2019centernet} to get the 3D box as we regress to a shared map  $\hat{S} \in R^{w \times h \times 2}$ storing the shared map width and height at the center location.

\begin{table*}[ht]
\begin{center}
\scriptsize
\begin{tabular}{lrrrrrrrrrrrr} 
\hline
\textbf{Method} &\textbf{mAP} &\textbf{NDS} & \textbf{Car} & \textbf{Truck}  & \textbf{Bus} & \textbf{Trailer} & \textbf{CV} & \textbf{Ped} & \textbf{Motor} & \textbf{Bicycle} & \textbf{TC} & \textbf{Barrier}\\
\hline\hline
FMF-PointPillars-Base (Ours) & 53.8  & 62.7 & \textbf{89.9} & 57.0 & 65.3 & 66.5 & 25.7 & 82.2 & 50.5 & 21.9 & 72.7 & 73.3  \\
CenterPoint \cite{yin2021centerbased} & 58.0 & 65.5 & 84.6 & 51.0 & 60.2 & 53.2  &17.5& 83.4 & 53.7& 28.7& 76.7& 70.9  \\
FMF-VoxelNet-Base (Ours) & \textbf{58.0} & \textbf{65.9} & 89.8 & \textbf{59.3} & \textbf{67.0} & \textbf{68.6} & \textbf{32.8} & \textbf{85.7} & \textbf{58.6} & \textbf{30.2} &   \textbf{77.4} & \textbf{77.0} \\
\hline
\end{tabular}
\end{center}
\caption{Comparison of the state-of-the-art methods  for 3D detection on nuScenes test set, Abbreviations: construction vehicle (CV), pedestrian (Ped), motorcycle (Motor), and traffic cone (TC).}
\label{tab:testset}
\end{table*}

\begin{table*}[ht]
\begin{center}
\scriptsize
\begin{tabular}{lrrrrrrrrr}
\hline
\textbf{Method} &\textbf{mAP↑} &\textbf{mATE↓} &\textbf{mASE↓} &\textbf{mAOE↓} &\textbf{mAVE↓} &\textbf{mAAE↓} &\textbf{NDS↑} & \\
\hline\hline
CenterPoint-PointPillars &0.5024 &0.3130 &0.2593 &0.3936 &\textbf{0.3260} &\textbf{0.1976} &0.6023 & \\
FMF-PointPillars-Base &\textbf{0.5256} &\textbf{0.3003} &\textbf{0.2564} & \textbf{0.3709} &0.3267 &0.1978 & \textbf{0.6176} & \\
\hline
Centerpoint-VoxelNet	& 0.5543& 0.3035& 0.2591& \textbf{0.3223}& 0.3096& 0.1944 & 0.6283& \\
FMF-VoxelNet-Base & \textbf{0.5719} &	\textbf{0.2964}	& \textbf{0.2552} &	0.3258 &	\textbf{0.2793} &	\textbf{0.1860} &	\bf{0.6517} \\
\hline
\end{tabular}
\end{center}
\caption{Ablation studies for 3D object detection on nuScenes validation set.}
\label{tab:val}
\end{table*}

\begin{table}[ht]
\begin{center}
\scriptsize
\begin{tabular}{lrrrrr}
\hline
\textbf{Method} &\textbf{Veh↑} & \textbf{Ped↑} &\textbf{Cyc↑} &\textbf{mAPH↑} & \textbf{Latency, ms}\\ 
\hline\hline
FMF-PointPllars (FP16)  & 69.65 & 54.61 & 62.28 & 62.18 & \textbf{62.30} \\  
FMF-VoxelNet &\textbf{70.74}&\textbf{65.46} &\textbf{67.63} &\textbf{67.95}  & 86.11  \\ 		
\hline
\end{tabular}
\end{center}
\caption{FMF modules performance for 3D detection on Waymo test set. We show the per-class and average Level 2 mAPH. Latency is shown for Tesla V100 GPU.}
\label{tab:way_tst}
\end{table}

\begin{table}[ht]
\begin{center}
\scriptsize
\begin{tabular}{lrrrrr}
\hline
\textbf{Method} &\textbf{Veh↑} & \textbf{Ped↑} &\textbf{Cyc↑} &\textbf{mAPH↑}& \textbf{Latency, ms}\\ 
\hline\hline 
CenterPoint-PointPillars & \textbf{65.5}	& 55.1 & 60.2 &	60.3 & \textbf{52.83} \\
FMF-PointPillars(FP16) & 61.75 & 58.12 & 65.26 & 62.19 & 62.30  \\ 
FMF-PointPillars  & 62.35 & \textbf{59.67} & \textbf{66.71} & \textbf{62.43} & 82.06 \\ 
\hline
CenterPoint-VoxelNet  & 66.2 & 62.6 & 67.1 &	64.7 & 84.05 \\  
FMF-VoxelNet(FP16) &66.25&63.06 &67.24 &65.85  & \textbf{77.53}\\
FMF-VoxelNet &\textbf{66.68}&\textbf{63.62} &\textbf{68.64} &\textbf{65.98} & 86.11 \\ 		
\hline
\end{tabular}
\end{center}
\caption{Comparison between one-stage CenterPoint and FMF-based methods for 3D detection on Waymo validation. We show the per-class and average Level 2 mAPH. Latency is shown for Tesla V100 GPU.}
\label{tab:way_val}
\end{table}

We used rendered Gaussian kernels at the object center $q_i$ for each class $c_i \in \{1 ...K\}$ and the target heat map for a pixel $p$ is 
$M_{p, k}=\max _{i: c_{i}=k} \exp \left(-\frac{\left(p-q_{i}\right)^{2}}{2 \sigma_{i}^{2}}\right)$, where
$\sigma_{i}$ is related to the size of the object.
\begin{equation}
L_{h m}=-\frac{1}{N} \sum_{p, k}\left\{\begin{array}{ll}
\left(1-z\right)^{\alpha} \log \left(z\right), \text{if } y=1, \\
\left(1-y\right)^{\beta}\left(z\right)^{\alpha} \log \left(1-z\right) \text {, o/w,}
\end{array}\right.
\end{equation} where $z = \hat{M}_{p, k}$ is resulting center coordinate heat map for $k$-th class and $y=M_{p, k}$, 'o/w' is abbreviation for 'otherwise'.

We use $L_{1}$ loss as 3D box size prediction loss $L_{s}$ at the Ground Truth centers and regress to a local offset, supervised by $L_{l}$ loss to minimize stride-errors as in \cite{10.1007/978-3-030-58589-1_5}. Height ($L_{H}$), rotation angle ($L_{r}$) and velocity ($L_{v}$) attributes are also trained with $L_{1}$ loss.
Training process is supervised by a total loss $L$ combining the all mentioned above losses:
\begin{equation}
    L = L_{hm}+\gamma_{l}L_{l}+\gamma_{s}L_{s}+\gamma_{H}L_{H}+\gamma_{r}L_{r}+\gamma_{v}L_{v},
\end{equation}
where 
$\gamma_{i}$ is the loss weight for the corresponding attribute, $\gamma_{l}=\gamma_{s}=\gamma_{H}=\gamma_{v}=1.0$ and $\gamma_{r}=0.2$.

\section{Experiments}
We mainly evaluated FMFNet on nuScenes and Waymo datasets. We implemented FMFNet using two common 3D encoders for point clouds: VoxelNet \cite{zhou2017voxelnet} and PointPillars \cite{8954311}.

\subsection{Implementation Details}
First, a voxelization setup is applied to sample the point cloud randomly into pillars or voxels depending on the used model. We sampled up to 60,000 pillars with a pillar size of [0.32, 0.32, 6.0] meters and up to 150,000 voxels with voxel size of [0.1, 0.1, 0.15] meters. The maximum number of points in the pillar is 20 while the voxel could contain up to 10 points. We fed these pillars/voxels into the backbone. We followed \cite{8954311} for PointPillars backbone and \cite{zhou2017voxelnet} for VoxelNet backbone.

The network is trained by AdamW optimizer with one-cycle learning rate policy \cite{loshchilov2019decoupled} with an initial learning rate 0.003 and batch size 4 equally distributed on 2 Tesla V100 GPU cards with weight decay 0.01, and momentum 0.85 to 0.95.
We trained our models for 32 epochs. We adopted data augmentation strategy in order to prevent overfitting. We used random flipping along both $X$ and $Y$ axis, and global scaling with a random factor from [0.95; 1.05] with a random global rotation between $[-\pi / 8, \pi / 8]$.


\subsection{Performance}
We compared the performance of our main 3D detection models with the state-of-the-art methods on nuScenes dataset. Our implemented FMF-PointPillars models run in real-time at 19FPS on GeForce RTX™ 3060 Ti while the most accurate FMF-VoxelNet model runs at 14FPS.
Table \ref{tab:testset} shows mean average precision (mAP), nuScenes detection score (NDS) and the mAP for each category on nuScenes test set. Table \ref{tab:way_tst} shows Level 2 per-class precision and Level 2 mean average precision with heading on Waymo 3D detection test set. Our FMF-VoxelNet model achieves a better performance in all 10 nuScenes categories and 3 Waymo categories compared to the State-of-the-art baseline method, which make FMF-PointPillars a reliable and strong baseline for a real-time 3D object detection pipeline from LiDAR point clouds. 

\subsection{Ablation Study}
All ablation studies are conducted on nuScenes validation dataset~\cite{caesar2020nuscenes}.

We first compared our detection approach results with the state-of-the-art center-based 3D object detection and tracking method \cite{yin2021centerbased} (our baseline). Our main 3D object detection results on nuScenes validation set are shown in Table \ref{tab:val} and on Waymo validation set are shown in Table~\ref{tab:way_val}.
Our FMF detectors perform much better than the baseline when category objects are rare and small, reflecting the model’s ability to capture the rotation and size invariance when detecting objects. These results convincingly highlight the advantage of using FMF approach. 


Our context based aggregation approach achieves similar results compared to FMF-block with shared convolutions layers on the validation set.

\section{Conclusion}
In this work, we have proposed feature map flow module to improve quality of center-based deep neural network architecture. The FMF-VoxelNet-Base model achieved better performance on the nuScenes an Waymo benchmark for real-time 3D detection.

{\small
\bibliographystyle{ieee_fullname}
\bibliography{fmf}
}

\end{document}